\pdfoutput=1

\documentclass[11pt]{article}

\usepackage[final]{acl}

\usepackage{times}
\usepackage{latexsym}
\usepackage{subcaption}
\usepackage{amsmath}
\usepackage{amssymb}
\usepackage{pifont}
\usepackage{booktabs}
\usepackage{multirow}
\usepackage{makecell}
\newcommand{\cmark}{\checkmark}
\newcommand{\xmark}{\ding{55}}

\usepackage[T1]{fontenc}

\usepackage[utf8]{inputenc}

\usepackage{microtype}

\usepackage{inconsolata}

\usepackage{graphicx}

\title{Editing Across Languages: \\ A Survey of Multilingual Knowledge Editing}

\author{
Nadir Durrani\thanks{The authors contributed equally} \hspace{11mm} Basel Mousi\footnotemark[1] \hspace{11mm} Fahim Dalvi \\
Qatar Computing Research Institute, HBKU, Doha, Qatar \\ 
{\tt \{ndurrani,bmousi,faimaduddin\}@hbku.edu.qa} \\ 
}

\begin{document}
\maketitle
 
\begin{abstract}

While Knowledge Editing has been extensively studied in monolingual settings, it remains underexplored in multilingual contexts. This survey systematizes recent research on \textbf{\textit{Multilingual Knowledge Editing (MKE)}}, a growing subdomain of model editing focused on ensuring factual edits generalize reliably across languages. We present a comprehensive taxonomy of MKE methods, covering parameter-based, memory-based, fine-tuning, and hypernetwork approaches.  We survey available benchmarks, summarize key findings on method effectiveness and transfer patterns, and identify persistent challenges such as cross-lingual propagation, language anisotropy, and limited evaluation for low-resource and culturally specific languages. We also discuss broader concerns such as stability and scalability of multilingual edits. Our analysis consolidates a rapidly evolving area and lays the groundwork for future progress in editable language-aware LLMs.

\end{abstract}

\section{Introduction}


The static nature of pre-trained LLMs presents an important limitation: factual knowledge inevitably becomes outdated or incorrect over time. To address this, the research community has developed knowledge editing (KE) techniques \cite{KnowledgeEditor,mitchell2022memorybasedmodeleditingscale,ROME} that aim to modify specific factual information within LLMs efficiently, without costly full retraining or broad performance degradation.

While KE methods have made impressive progress in monolingual settings, they overlook a critical real-world dimension: \textbf{multilinguality}. Multilingual LLMs are increasingly deployed in cross-lingual applications, and it is insufficient for a knowledge update to take effect only in a single language. Ideally, an edit in one language should propagate across all supported languages, ensuring consistent and equitable access to updated knowledge regardless of linguistic background. This challenge has given rise to the emerging field of \textbf{Multilingual Knowledge Editing (MKE)}, which seeks to synchronize factual updates across languages while preserving unrelated knowledge. 


In this paper, we present the \textbf{first comprehensive survey of methods and benchmarks for multilingual knowledge editing}. We review recent advancements in (i) proposed MKE methodologies, including parameter editing methods, memory-based methods, fine-tuning methods, and hypernetwork-based methods; (ii) benchmark datasets designed to evaluate multilingual knowledge editing effectiveness and cross-lingual generalization; (iii) empirical findings on method effectiveness and cross-lingual generalization; and (iv) key open challenges and future directions toward building LLMs that are reliably editable across languages. Our goal is to provide a structured foundation for future research and help unify this fast-growing but currently scattered domain.

While several surveys have examined knowledge editing in LLMs, they focus primarily on monolingual settings or broad architectural taxonomies. \citet{mazzia2024surveyknowledgeeditingneural} provide a general overview of neural model editing across modalities. \citet{Song-Survey-24} emphasize technical classifications and optimization formulations, while \citet{zhang-Survey-2024} focus on monolingual editing and interpretability. None of these surveys consider the specific challenges of multilingual propagation or evaluation.

By contrast, our survey focuses explicitly on the multilingual dimension. We contribute (i) a multilingual reinterpretation of existing method families, (ii) a synthesis of empirical trends observed across MKE studies, and (iii) targeted visualizations comparing methods, languages, and model scales. Our work complements and extends prior surveys by highlighting multilingual knowledge editing as a critical but underexplored area.

\section{Multilingual Knowledge Editing}
\label{sec:problem}

Large Language Models (LLMs) are parameterized functions \( f(x; \theta) \), where \( \theta \) denotes the model weights. While LLMs acquire extensive factual knowledge during pre-training, this knowledge remains fixed and may become outdated or incorrect.

\textbf{Knowledge Editing (KE)} modifies a model’s behavior to reflect updated facts, either by changing its parameters \( \theta \) or by influencing outputs through prompts or external memory. The objective is to apply edits locally, affecting only the intended fact while preserving unrelated knowledge.

This survey focuses on \textbf{Multilingual Knowledge Editing (MKE)}, where edits introduced in one language (e.g., English) should generalize to semantically equivalent queries in other languages, without disrupting unrelated outputs.

Let \( \mathcal{L} = \{\ell_1, \ell_2, \ldots\} \) denote the set of languages, and represent a fact as a triple \( (s, r, o_{\text{new}}) \). Given queries \( Q_{\ell_s} \) in a source language \( \ell_s \), and corresponding queries \( Q_{\ell_t} \) in target languages \( \ell_t \in \mathcal{L} \), the goal is to find updated parameters \( \theta' \)\footnote{Non-parametric methods apply edits at inference time through prompts or external memory without modifying \( \theta \).} such that:
\vspace{-2mm}
\[
f(q; \theta') \approx o_{\text{new}} \quad \text{for } q \in Q_{\ell_s} \cup Q_{\ell_t},
\]
\[
f(q; \theta') \approx f(q; \theta) \quad \text{for } q \notin Q_{\ell_s} \cup Q_{\ell_t}.
\]


A key challenge in MKE is \textit{language anisotropy}, where multilingual models encode different languages in misaligned subspaces, making edits difficult to propagate reliably. In this survey, we define the multilingual knowledge editing problem and highlight its core challenges. The following sections cover proposed MKE methods, benchmark datasets, empirical findings across models and languages, and open research directions.




\section{Methods in Multilingual KE}
\label{sec:methods}

Work in multilingual knowledge editing (MKE) builds on monolingual techniques and introduces new methods to support cross-lingual consistency. These approaches can be broadly grouped into four methodological families: (i) \textbf{Parameter Editing}, which modifies model weights directly; (ii) \textbf{Memory-based} methods, which use in-context learning or external memory without updating parameters; (iii) \textbf{Fine-tuning}, including full and instruction-tuned updates; and (iv) \textbf{Hypernetwork-based}, which generate edits via auxiliary networks. This taxonomy supports a structured comparison of design choices and trade-offs in MKE.


\subsection{Parameter Editing Methods}
\label{subsec:PEM}


Parameter editing methods modify the internal weights of LLMs to update factual knowledge.
They can be broadly divided into two categories: general methods originally developed for monolingual settings and later applied to multilingual models, and specialized methods designed to support cross-lingual propagation of edits.

\subsubsection{General KE Methods Adapted for MKE}

\textbf{ROME}~\cite{ROME} introduced the locate-then-edit paradigm, identifying key activation pathways and applying a low-rank update to model weights. 
\textbf{MEMIT}~\cite{MEMIT} extended ROME to support batch editing of multiple facts using rank-one updates \( \{\Delta \theta_k\}_{k=1}^{K} \) distributed across layers. While it preserves locality and improves efficiency, its effectiveness across languages is limited by language-specific neuron activations. Similarly, \textbf{KnowledgeNeuron}~\cite{KN} and \textbf{PMET}~\cite{PMET} followed the locate-and-edit paradigm by optimizing \( \Delta \theta \) to encode factual edits. 
\citet{wang-etal-2024-cross,zhang-etal-2025-multilingual} adapted \textbf{ROME}, \textbf{MEMIT}, and \textbf{PMET} to multilingual models. While these adaptations preserve edit locality and maintain computational efficiency, they suffer from poor cross-lingual propagation due to language-specific activation patterns and representational anisotropy.

\subsubsection{Specific Methods Designed for MKE}

\textbf{MEMAT}~\cite{MEMAT}  incorporated attention head analysis into the locate-then-edit paradigm to improve cross-lingual generalization. It ranks heads based on their contribution to predictions in target languages and applies targeted corrections.
\textbf{MPN}~\cite{si2024mpnleveragingmultilingualpatch} introduced patch neurons, key-value units trained to jointly encode facts across languages. These patches update knowledge without modifying original weights, promoting multilingual consistency. \textbf{LU-LAFNs}~\cite{zhang-etal-2025-multilingual} takes a different approach by identifying shared high-attribution neurons across languages and caching their values to apply edits dynamically at inference, enabling non-destructive, language-independent interventions. Finally, \textbf{MEMLA}~\cite{xie2024memlaenhancingmultilingualknowledge} used integrated gradients to locate language-relevant neurons and applies masked LoRA updates to targeted subsets, balancing cross-lingual generalization and edit locality.

\subsubsection{Summary}

\textbf{Strengths:} Parameter editing methods offer localized updates that reduce interference with unrelated knowledge and are efficient for factual edits without full retraining. They are also compatible with standard transformer architectures, making them easy to integrate into multilingual LLM pipelines.

\noindent \textbf{Weaknesses:} These methods rely on shared internal representations across languages, which are often weak or inconsistent. As a result, they do not guarantee reliable cross-lingual propagation; edits made in one language may fail to transfer to others. Their performance is also sensitive to tokenization and neuron alignment, particularly in morphologically rich or low-resource languages. While MKE-specific approaches (e.g LU-LAFNs) address these issues through attention or neuron-level interventions, key theoretical challenges remain.

\subsection{Memory-based Methods}
\label{subsec:MBM}


Memory-based methods perform knowledge editing without modifying model weights \( \theta \), using either in-context interventions or external memory to override factual outputs. These approaches offer greater flexibility and reduce the risk of unintended interference. We group them into two groups: \textbf{In-context Editing} and \textbf{Memory-Retriever} methods.

\subsubsection{In-Context Editing (ICE \& IKE)}

In-context editing~\cite{zheng-etal-2023-edit} modifies model behavior by injecting factual corrections into the input prompt, without changing model weights. Zero-shot ICE uses single factual statements, while IKE extends this with few-shot demonstrations. 
These methods can generalize across languages if equivalent demonstrations are available in each language. ~\citet{nie2025bmike53investigatingcrosslingualknowledge} for example applied in-context editing in 53 languages. 

However, these methods are constrained by context length: edits must be carried in the prompt, which reduces space for task inputs and outputs. This issue is especially severe in languages with high tokenizer fertility, where facts occupy more tokens, and in sequential editing scenarios, where multiple edits quickly accumulate. 

\subsubsection{Memory-Retriever Editing}

Memory-Retriever methods attach an external memory to the LLM and retrieve language-specific facts \( \{m_i\} \) at inference time, producing \( f(q, \{m_i\}; \theta) \approx o_{\text{new}} \) without modifying weights. This design supports multilingual edits by isolating updates outside the model, but effectiveness depends on retriever quality. Variants differ in how memory is used: \textbf{SERAC}~\cite{mitchell2022memorybasedmodeleditingscale} overrides predictions with matched entries. \textbf{ReMaKE}~\cite{wang-etal-2024-retrieval} uses retrieved memories to guide lightweight updates. \textbf{MQA-KEAL}~\cite{ali-etal-2025-mqa} performs multi-hop reasoning over memory chains. Cross-lingual retrievers like \textbf{Mello-CL}~\cite{khandelwal-etal-2024-cross} boost recall in multilingual and low-resource contexts.\footnote{Other approaches such as \textbf{GRACE}~\cite{GRACE} and \textbf{WISE}~\cite{WISE} also support handling multiple edits while maintaining locality. We did not include them in our main taxonomy since they were developed and evaluated only in English and do not address cross-lingual generalization, but they represent promising directions for future MKE research.}

\subsubsection{Summary}

\textbf{Strengths:} Memory-based methods avoid parameter updates, minimizing interference with unrelated cross-lingual knowledge. Their edits are applied externally at inference time, enabling high flexibility for dynamic and reversible updates. Because corrections are decoupled from internal weights, these approaches can be applied independently to any language, making them naturally suitable for multilingual deployments.

\noindent \textbf{Weaknesses:} Their effectiveness depends on language-specific components: prompt understanding and instruction following for in-context methods, or retriever accuracy and memory quality for retrieval-based approaches. Neither guarantees cross-lingual propagation; corrections must be explicitly provided or retrieved in each target language. In addition, memory-retriever methods incur storage and latency costs as edits accumulate, while in-context editing requires constructing effective prompts across languages.




\subsection{Fine-Tuning Methods}
\label{subsec:FTM}

Fine-tuning updates large parameter blocks, such as full weights or adapters, to encode factual edits. Unlike parameter editing, which operates on localized parameter subspaces, it offers stronger reliability but higher risk of side effects in multilingual settings. These methods fall into two types: \textit{baseline fine-tuning}, which trains directly on factual updates, and \textit{instruction-tuned fine-tuning}, which treats editing as a task via structured instructions.

\subsubsection{Baseline Fine-Tuning Methods}

Baseline fine-tuning methods apply supervised learning to update model parameters with edited knowledge. \textbf{Full Fine-Tuning (FT)} updates all model parameters \( \theta \) by minimizing loss over edited facts, making it a simple and commonly used baseline. \textbf{LoRA-FT}~\cite{LoRAFT} introduces low-rank adaptation matrices into transformer blocks and updates only these, offering a parameter-efficient alternative. \textbf{FT-L} constrains updates to selected layers using gradient masking~\cite{modifyingmemories}, while \textbf{FT-M} fine-tunes a specific MLP sub-layer. These methods, originally developed for monolingual editing, are applied as baselines (e.g. in ~\citet{XKDE}) for multilingual knowledge editing. While they offer strong edit reliability in the source language, generalization to target languages remains purely implicit, with no mechanism to enforce cross-lingual propagation.

\subsubsection{Instruction-Tuned Knowledge Editing}

Instruction-tuned methods frame editing as a task, training models to apply factual updates through structured edit instructions rather than direct parameter overwriting. They are fine-tuned on supervised pairs of edit requests and desired outputs, enabling the model to internalize editing behavior.

\textbf{Learning to Edit (LTE)}~\cite{LTE} generates synthetic edits and fine-tunes the model to process an edit descriptor \( E \) and query \( q \) to output:
\vspace{-4mm}
\[
f(q, E; \theta') \approx o_{\text{new}} \quad \forall q \in Q_{\ell_s}.
\]

\textbf{X-KDE}~\cite{XKDE} extended LTE to multilingual settings via a two-stage framework: Cross-lingual Edition Instruction Tuning (XE-IT) trains on parallel data to transfer edits across languages, and Target-Language Preference Optimization (TL-PO) aligns outputs for consistency.
\vspace{-2mm}
\[
f(q, E; \theta') \approx o_{\text{new}} \quad \forall q \in Q_{\ell_s} \cup Q_{\ell_t}, \, \ell_t \in \mathcal{L}.
\]

Instruction-tuned approaches are the first systematic attempt to model editing as a generalizable capability. LTE is effective in monolingual contexts, while X-KDE introduced a supervised framework for cross-lingual propagation and alignment.

\subsubsection{Summary}

\textbf{Strengths:} Fine-tuning methods are widely used for knowledge editing due to their simplicity and strong factual reliability. They are compatible with pre-trained transformer architectures and require minimal structural changes. LoRA-based fine-tuning improves efficiency by updating a small parameter subset. Instruction-tuned approaches, such as LTE and X-KDE, treat editing as a learnable task, offering more control over behavior and reducing unintended side effects. Notably, X-KDE is the first supervised fine-tuning framework specifically designed to promote cross-lingual consistency.

\noindent \textbf{Weaknesses:} Fine-tuning provides no guarantees for cross-lingual propagation; generalization to target languages depends on implicit internal alignment. Large parameter updates also risk altering unrelated knowledge. While instruction-tuned methods mitigate this by explicitly modeling the editing process, they rely on large, curated datasets and careful instruction design.

\subsection{Hypernetwork-Based Methods}
\label{subsec:hypernetwork}


Hypernetwork-based methods edit knowledge by training  auxiliary networks to generate updates. Instead of manual weight changes, hypernetwork learns to produce edits from query-output pairs. \textbf{MEND}~\cite{MEND} introduced this by training a hypernetwork \( h_{\phi} \) to produce updates \( \Delta \theta = h_{\phi}(q, o_{\text{new}}) \) such that \( f(q; \theta + \Delta \theta) \approx o_{\text{new}} \), while minimizing interference. \textbf{KnowledgeEditor}~\cite{KnowledgeEditor} added regularization to stabilize edits and improve generalization. 
\textbf{LiME}~\cite{LIME} adapted this paradigm for multilingual editing by conditioning the hypernetwork on source language \( \ell_s \), generating edits in language-specific subspaces: \( f(q; \theta + h_{\phi}(q, o_{\text{new}}, \ell_s)) \approx o_{\text{new}} \). By modeling language anisotropy explicitly, LiME improved cross-lingual stability.

\subsubsection{Summary}

\textbf{Strengths:} Hypernetwork-based methods enable meta-learning of edits without manual parameter selection, offering fast, localized updates with minimal disruption. Language-conditioned variants such as LiME introduce explicit mechanisms to improve edit stability across languages.

\noindent \textbf{Weaknesses:} The effectiveness of hypernetwork-based methods depends heavily on the quality and coverage of training data. While \textbf{MEND} and \textbf{KnowledgeEditor} were designed for monolingual models and used as multilingual baselines, \newcite{beniwal-etal-2024-cross} showed their limited cross-lingual transfer. \textbf{LiME} mitigates this by conditioning on the source language, but its reliance on language-specific subspaces may limit generalization across distant language pairs.

\subsection{Summary of Methods}

\begin{table}[t]
\centering
\resizebox{\columnwidth}{!}{
\begin{tabular}{l c c c c c}
\toprule
\textbf{Method} & \textbf{Edit Rel} & \textbf{X-ling Prop} & \textbf{Locality} & \textbf{Param. Mod} & \textbf{Dyn. Flex} \\
\midrule
\multicolumn{6}{c}{\textbf{Parameter Editing Methods}} \\
\midrule
ROME & \cmark & \xmark & \cmark & \cmark & \xmark \\
MEMIT & \cmark & \xmark & \cmark & \cmark & \xmark \\
MEMAT & \cmark & $\sim$ & \cmark & \cmark & \xmark \\
MPN & \cmark & $\sim$ & \cmark & \cmark & \xmark \\
LU-LAFNs & \cmark & $\sim$ & \cmark & \xmark & \cmark \\
MEMLA & \cmark & $\sim$ & \cmark & \cmark & \xmark \\
\midrule
\multicolumn{6}{c}{\textbf{Memory-Based Methods}} \\
\midrule
IKE / ICE & $\sim$ & $\sim$ & \cmark & \xmark & \cmark \\
SERAC & \cmark & \xmark & \cmark & \xmark & \cmark \\
ReMaKE & \cmark & $\sim$ & \cmark & \xmark & \cmark \\
MQA-KEAL & \cmark & \xmark & \cmark & \xmark & \cmark \\
MELLO-CL & $\sim$ & $\sim$ & \cmark & \xmark & \cmark \\
\midrule
\multicolumn{6}{c}{\textbf{Fine-Tuning Methods}} \\
\midrule
FT & \cmark & \xmark & \xmark & \cmark & \xmark \\
LoRA-FT & \cmark & \xmark & $\sim$ & \cmark & \xmark \\
FT-L / FT-M & \cmark & \xmark & $\sim$ & \cmark & \xmark \\
LTE & \cmark & \xmark & $\sim$ & \cmark & \xmark \\
X-KDE & \cmark & $\sim$ & $\sim$ & \cmark & \xmark \\
\midrule
\multicolumn{6}{c}{\textbf{Hypernetwork-Based Methods}} \\
\midrule
MEND & \cmark & \xmark & \cmark & \cmark & \xmark \\
KnowledgeEditor & \cmark & \xmark & \cmark & \cmark & \xmark \\
LiME & \cmark & $\sim$ & \cmark & \cmark & \xmark \\
\bottomrule
\end{tabular}
}
\caption{Comparison of multilingual knowledge editing methods across core evaluation criteria. \textbf{Edit Rel}: edit reliability in source language; \textbf{X-ling Prop}: cross-lingual propagation; \textbf{Locality}: preservation of unrelated knowledge; \textbf{Param. Mod}: modifies model weights; \textbf{Dyn. Flex}: supports inference-time edit control. \cmark = strong; $\sim$ = partial; \xmark = absent.}
\label{tab:mke_comparison}
\vspace{-2mm}
\end{table}

\noindent
Table~\ref{tab:mke_comparison} summarizes the trade-offs across multilingual knowledge editing methods. Parameter editors offer strong locality and reliability but struggle with propagation. Memory-based methods are flexible and reversible, yet rely on language-specific retrieval or prompting. Fine-tuning provides reliable edits but lacks adaptability. Instruction-tuned approaches such as X-KDE improve generalization with explicit supervision. Hypernetwork methods like LiME minimize interference but remain underexplored in multilingual contexts. No method meets all requirements, motivating further research into hybrid strategies.

\begin{table}[t]
\centering
\resizebox{\columnwidth}{!}{
\begin{tabular}{l c c p{3.6cm}}
\toprule
\textbf{Benchmark} & \textbf{Langs} & \textbf{Human?} & \textbf{Notes} \\
\midrule
MLaKE~\cite{wei-etal-2025-mlake} & 5 & \cmark & ChatGPT-generated, human-verified; no paraphrase/locality samples \\
MzSRE~\cite{wang-etal-2024-retrieval} & 12 & \xmark & Translated via Google; used for all four metrics \\
biZsRE~\cite{wang-etal-2024-cross} & 2 & $\sim$ & GPT-translated; test set verified by native speakers \\
BMIKE-53~\cite{nie2025bmike53investigatingcrosslingualknowledge} & 53 & $\sim$ & GPT4o-translated; BLEU + manual inspection \\
MQuAKE~\cite{khandelwal-etal-2024-cross} & 7 & \cmark & Translations verified via BLEU and human checks \\
MQuAKE-AR~\cite{ali-etal-2025-mqa} & 1 & \cmark & LLM-based scoring with manual refinement \\
MKEB~\cite{xie2024memlaenhancingmultilingualknowledge} & 12 & \xmark & SPARQL + Baidu; no human revision \\
DocTer~\cite{wu2023evakellmnewbenchmarkevaluating} & 2 & ? & Generated from Counterfact using ChatGPT \\
Counterfact (CA)~\cite{MEMAT} & 2 & \cmark & Aina-translated; SimAlign + human filtering \\
CKnowEdit~\cite{fang2024cknowedit} & 1 & \cmark & Chinese cultural and linguistic error correction \\
\bottomrule
\end{tabular}
}
\caption{Multilingual knowledge editing benchmarks. Langs. = number of languages. Human? = human annotation or translation verification.}
\vspace{-4mm}

\label{tab:mke_benchmarks}
\end{table}


\section{Evaluation for Multilingual KE}
\label{sec:benchmarking}

Despite recent progress in multilingual knowledge editing (MKE), the field still lacks a unified evaluation framework. Most studies build on protocols developed in the monolingual KE literature, typically using four core criteria:

\textbf{Reliability} measures whether the edit is successful in the source language, typically quantified as the percentage of source-language queries that yield the updated fact.

\textbf{Generality} evaluates robustness to linguistic variation, assessing model accuracy on paraphrased or rephrased queries in the source language.

\textbf{Locality} examines how unrelated knowledge is preserved, usually by comparing model predictions on non-target queries before and after the edit.

\textbf{Portability} captures the extent to which an edit generalizes beyond its original setting, including transfer to downstream tasks, multi-hop reasoning chains, or different model variants.

In the multilingual setting, these dimensions are extended to account for \textit{cross-lingual propagation}, whether an edit made in one language transfers accurately to semantically equivalent queries in other languages. This multilingual extension has become a defining challenge in MKE, and features prominently in recent work~\cite{wei-etal-2025-mlake, zhang-etal-2025-multilingual, XKDE, xie2024memlaenhancingmultilingualknowledge}.\footnote{Terminology varies across studies. For example, \citet{xie2024memlaenhancingmultilingualknowledge} refer to reliability as the efficacy score, generality as the paraphrase score, and locality as the neighborhood score.}

Some recent work~\cite{ali-etal-2025-mqa, khandelwal-etal-2024-cross} further expands evaluation by introducing \textit{multi-hop reasoning} over edited knowledge, using variants of the multi-hop accuracy metric introduced by~\citet{zhong-etal-2023-mquake}. These setups test whether edits are faithfully integrated into the model’s broader knowledge base and whether they support multilingual inference chains.

To support evaluation, a growing number of multilingual benchmarks have emerged. These datasets vary in language coverage, task complexity (e.g., single-hop vs. multi-hop), and translation quality, reflecting the diversity of MKE settings. Table~\ref{tab:mke_benchmarks} summarizes key characteristics. Additional details on construction and evaluation are in Appendix~\ref{appendix:benchmarks}.

\section{Findings}
\label{sec:analysis-findings}

We now highlight recurring trends and differences across method families, focusing on how these methods perform under the evaluation dimensions discussed in the previous section.

\begin{figure}[t]
    \centering
    \includegraphics[width=0.95\linewidth]{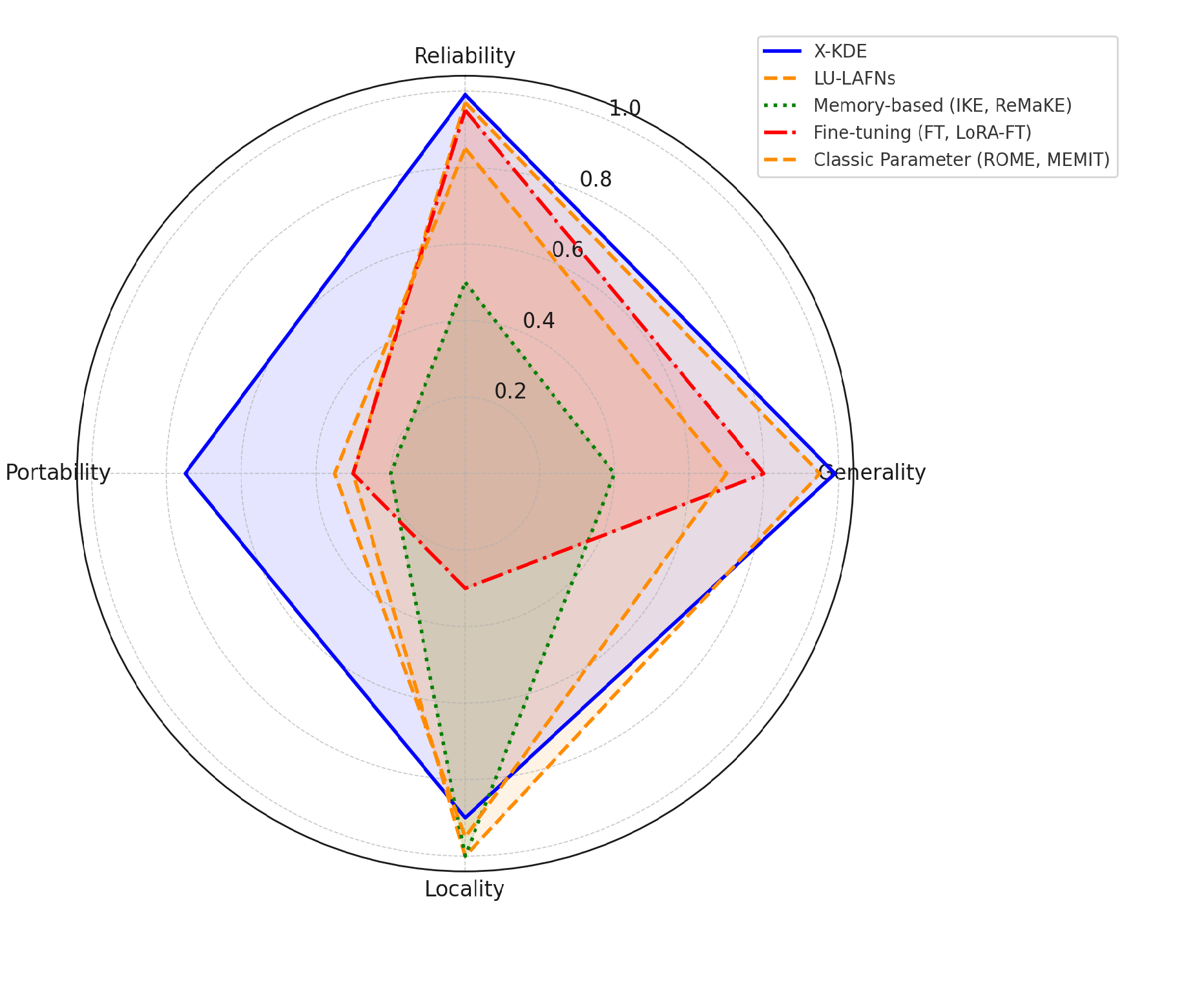}
    \vspace{-4mm}
   \caption{Comparison of multilingual knowledge editing methods across four evaluation criteria: reliability, generality, locality, and portability. Methods are grouped and color-coded by method family. Scores represent an approximate synthesis of reported results from recent studies ~\cite{zhang-etal-2025-multilingual,XKDE}}
    \label{fig:comparative_analysis}
    \vspace{-4mm}
\end{figure}

\subsection{Comparative Evaluation}

\textit{Q: Which methods best balance reliability, generality locality and portability in MKE?}

Experimental comparisons of multilingual knowledge editing (MKE) methods reveal patterns in how different approach families trade off reliability, generality, locality, and portability.

\paragraph{Parameter Editing} Parameter-based methods such as ROME and MEMIT perform strongly in monolingual settings but struggle in multilingual contexts, often causing interference and degraded cross-lingual transfer~\cite{wang-etal-2024-retrieval,XKDE,zhang-etal-2025-multilingual,wei-etal-2025-mlake,xie2024memlaenhancingmultilingualknowledge}. 
LU-LAFNs achieves perfect locality and strong reliability by identifying and modifying shared factual neurons across languages. A key reason for its superior locality is how updates are applied: after optimization, the updates are stored externally and only injected at inference time if a matching subject is detected in the user prompt. This minimizes the risk of unintended changes to unrelated knowledge. However, like other parameter editing methods, LU-LAFNs shows weaker cross-lingual generalization and portability.

\paragraph{Memory-based Methods} Memory-based methods like IKE and ReMaKE avoid parameter updates and instead rely on retrieval or prompting. ReMaKE achieves excellent locality but suffers from low reliability in many multilingual scenarios, with performance often dropping below 50\%. In contrast, IKE demonstrates relatively poor locality, as it heavily depends on the quality and selection of few-shot exemplars \cite{wang-etal-2024-cross, wang-etal-2024-retrieval, XKDE}. Additionally, the applicability of ReMaKE across model families remains limited, its performance degrades significantly when applied to models such as BLOOM~\cite{zhang-etal-2025-multilingual}, indicating strong dependence on the underlying model's capabilities. Despite their limitations, prompting-based approaches remain surprisingly competitive: \citet{nie2025bmike53investigatingcrosslingualknowledge} demonstrated that simple in-context editing achieves reasonable cross-lingual edit success across 53 languages, even without model weight modifications.

\paragraph{Fine-tuning} Fine-tuning strategies (e.g., FT-L, FT-M, LoRA-FT) deliver high reliability for monolingual edits but perform worst in locality, often degrading unrelated knowledge~\cite{LIME,XKDE}. They also exhibit limited ability to transfer edits across languages, making them the least favorable option for MKE in general. An important exception is X-KDE~\cite{XKDE}, which combines fine-tuning with cross-lingual instruction tuning and preference optimization to achieve the best reported overall balance of reliability, generality, and locality across languages. Among fine-tuning methods without explicit adaptation for MKE, LTE stands out for its strong cross-lingual generalization despite being trained only on monolingual synthetic edits~\cite{XKDE}.

\paragraph{Hypernetwork-based Methods} No large-scale multilingual benchmarking of hypernetwork-based methods exists. Early results in monolingual and small-scale multilingual settings suggest potential for efficient edits and strong generalization, though systematic evidence remains limited.

\paragraph{Summary} Current evidence consistently identifies \textbf{X-KDE as offering the best overall balance of reliability, generality, and locality across languages. LU-LAFNs stands out within parameter editing methods} for its unmatched locality through targeted neuron editing and conditional application of updates, although it exhibits weaker cross-lingual portability. Classic parameter editors and fine-tuning approaches perform substantially worse in cross-lingual settings, and memory-based methods, while safe in terms of locality, offer limited reliability and portability. Hypernetwork-based methods remain promising but underexplored in the context of multilingual knowledge editing. Figure~\ref{fig:comparative_analysis} provides a visual summary of the relative performance of these method families across the four evaluation criteria.

\begin{figure}[t]
    \centering
    \includegraphics[width=0.5\textwidth]{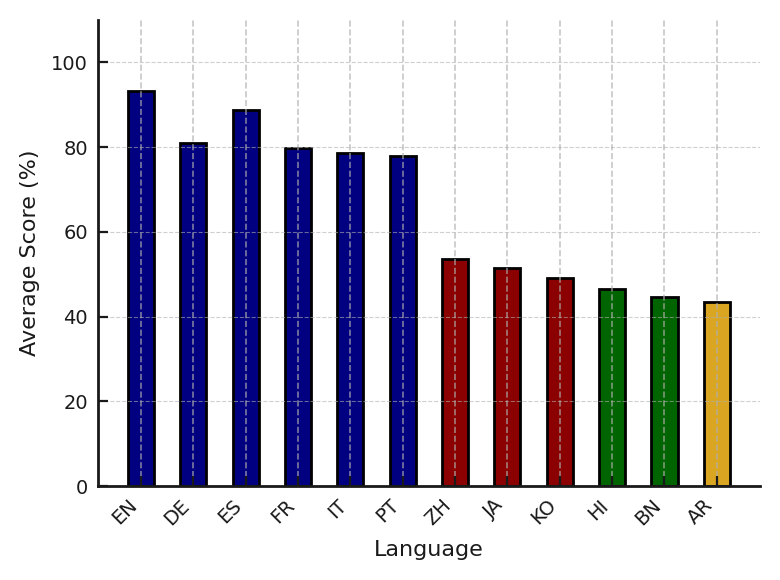}
    \caption{Average Reliability and Generality for MEMLA~\cite{xie2024memlaenhancingmultilingualknowledge} across twelve languages, illustrating a typical directionality pattern: edits in high-resource  or related languages transfer better than those in low-resource or distant ones.
}
    \label{fig:memla_directionality}
    \vspace{-4mm}
\end{figure}

\subsection{Cross-Lingual Transferability}

\textit{Q: What factors influence the cross-lingual propagation of edits in MKE, and how well do current methods address this challenge?}

Experimental studies on MKE consistently show that cross-lingual transfer remains a major challenge. Analysis of recent benchmarks highlights several key factors influencing the effectiveness of cross-lingual propagation of edits.

\paragraph{Language Relatedness} The strongest predictor of successful cross-lingual propagation is the degree of linguistic similarity between the source and target languages. Edits propagate better within the same language family (e.g., English to Spanish or German (see Figure~\ref{fig:memla_directionality})) than between distant language families (e.g., English to Chinese or Arabic)~\cite{mousiKE2025,zhang-etal-2025-multilingual,wei-etal-2025-mlake}. Some studies further suggest that shared writing systems or similar orthographic structures (e.g., Latin-based scripts) may offer marginal additional benefits to transferability~\cite{nie2025bmike53investigatingcrosslingualknowledge,LIME}. ~\citet{beniwal-etal-2024-cross} also showed that script differences (e.g., Devanagari vs. Tamil) reduce edit transferability, highlighting script divergence as a secondary but relevant limiting factor. Recent work further underscores the role of subject token similarity in transfer success: ~\citet{MEMAT} showed that edits are more likely to propagate when the embedding representations of subject entities are closely aligned across languages.

\paragraph{Language and Directionality of Transfer} The language in which the edit is applied has a notable impact on cross-lingual propagation. Edits performed in high-resource languages such as English generally transfer more effectively to other languages than edits made in lower-resource or non-Latin languages such as Arabic~\cite{mousiKE2025} or Chinese~\cite{XKDE,zhang-etal-2025-multilingual,xie2024memlaenhancingmultilingualknowledge,nie2025bmike53investigatingcrosslingualknowledge}. This asymmetry likely reflects the dominant role of English and other high-resource languages during pretraining and fine-tuning of multilingual language models.



\paragraph{Challenges in Multi-hop Reasoning} Transferability further degrades in multi-hop reasoning settings, where complex fact chains must be updated across multiple languages~\cite{wei-etal-2025-mlake,xie2024memlaenhancingmultilingualknowledge}. Existing methods show significant performance drops in multi-hop multilingual scenarios compared to single-hop edits.

\paragraph{Overall Limitations} Despite recent progress, no  method achieves fully consistent and reliable cross-lingual propagation across all language pairs. Multiple studies report large variability in performance, especially for distant and low-resource languages~\cite{nie2025bmike53investigatingcrosslingualknowledge,wei-etal-2025-mlake,xie2024memlaenhancingmultilingualknowledge}. Cross-lingual transferability thus remains an open challenge for the field.

\subsection{Impact of Model Size and Capability}

\begin{figure}[t]
    \centering
    \includegraphics[width=0.9\linewidth]{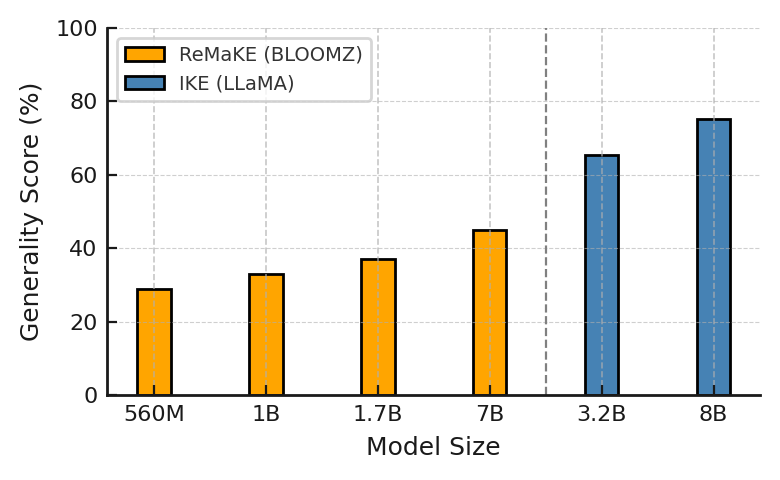}
    \caption{Generality scores for BLOOMZ and LLaMA models across increasing model sizes. ReMaKE~\cite{wang-etal-2024-retrieval} (BLOOMZ) results are extracted for mzsRE generality, averaged across 10 languages. IKE~\cite{nie2025bmike53investigatingcrosslingualknowledge} (LLaMA) results are extracted for mzsRE generality under zero-shot settings, averaged across 53 languages. Results show that generality consistently improves with model scale for both models.}
    \label{fig:model_comparison}
    \vspace{-2mm}
\end{figure}

\textit{Q: How do model size affect the performance of multilingual knowledge editing?}


Larger and more capable models consistently yield higher editing quality across reliability, generality, locality, and, to a lesser extent, portability. LLaMA-3.1-8B outperforms Qwen2-7B and Bloomz-7B, while Qwen2.5-7B-Instruct surpasses Chinese-LLaMA-7B in cross-lingual benchmarks~\cite{XKDE,zhang-etal-2025-multilingual}.

An interesting secondary finding is that stronger models also stabilize weaker editing methods. For example, fine-tuning performs poorly on weaker models like Bloomz but becomes stable, though still suboptimal on LLaMA-3 and Qwen2~\cite{zhang-etal-2025-multilingual}. Parameter editors like M-MEMIT show similar improvements on stronger models.


Instruction tuning further enhances robustness. Models such as Qwen2.5-7B-Instruct and LLaMA-chat exhibit slightly better generality and portability as shown in ~\citet{XKDE}. This suggests instruction-tuned models may possess a more disentangled and structured internal knowledge space, which aids editing consistency.

Finally, stronger models reduce the performance gap across methods. On Bloomz-7B, the performance difference between LU-LAFNs and M-MEMIT exceeds 25 points, whereas on LLaMA-3.1-8B the gap shrinks to under 10 points~\cite{zhang-etal-2025-multilingual}. However, the hardest remaining challenge is portability. Even strong models struggle to propagate edits across English and Chinese, with only X-KDE and LU-LAFNs maintaining acceptable performance in this setting~\cite{XKDE}.

\section{Discussion}
\label{sec:discussion}


Multilingual knowledge editing has made progress, but major limitations remain. In this section, we identify persistent challenges and outline opportunities for future work. We organize these into thematic groups, reflecting both empirical trends and conceptual gaps in the literature.

\subsection{Open Challenges}


\paragraph{Language Anisotropy and Representation Misalignment}
Multilingual LLMs encode languages into overlapping yet partially distinct internal subspaces \cite{mousi-etal-2024-exploring}, creating structural anisotropy that can hinder cross-lingual transfer. A key open problem for MKE is to bridge these spaces by using shared regions and handling language-specific pockets.

\paragraph{Lack of Evaluation Coverage for Low-Resource Languages}
Most MKE studies evaluate only high-resource languages (e.g., English, Chinese). Systematic evaluation on low-resource, morphologically complex, and under-represented languages (e.g., Swahili, Urdu) remains largely absent. In addition, current MKE benchmarks largely assume that factual knowledge is language-invariant. This overlooks culturally specific knowledge that may not have direct equivalents across languages, which is especially important for low-resource languages where such references are more prominent. Developing evaluation resources that account for cultural grounding remains an open challenge.

\paragraph{Trade-off Between Locality and Propagation}
Methods that enforce strong locality (e.g., LU-LAFNs) tend to sacrifice propagation, while those that encourage transfer (e.g., fine-tuning, prompting) risk damaging unrelated knowledge. Designing architectures or training objectives that mitigate this tension remains an open research direction.

\paragraph{Stability and Model Collapse} Another critical challenge, underexplored in multilingual settings, is model collapse, where even a few edits can substantially degrade an LLM’s general performance \cite{green-etal-2025-babeledits}. Monolingual studies, \cite{yang-etal-2024-butterfly,gupta-etal-2024-rebuilding} show that edits may trigger instability, and multilingual scenarios, with misaligned representational spaces, exacerbate this risk. Evaluating MKE methods should therefore consider both the success of edits and the preservation of model capabilities, particularly under sequential or batched updates where collapse risks accumulate.

\begin{table}[t]
\centering
\resizebox{\columnwidth}{!}{
\begin{tabular}{l c c l}
\toprule
\textbf{Method Family} & \textbf{Sequential} & \textbf{Batch} & \textbf{Notes} \\
\midrule
Parameter Editing 
& \makecell[l]{\xmark\ (ROME) \\ \cmark\ (LU-LAFNs) \\ \(\sim\)\ (MEMLA)} 
& \makecell[l]{\xmark\ (ROME) \\ \cmark\ (MEMIT, MEMLA, LU-LAFNs)} 
& \makecell[l]{MEMIT supports batch; \\ LU-LAFNs allow partial \\ sequential edits.} \\
\midrule
Memory-Based 
& \cmark 
& \cmark 
& \makecell[l]{Edits stored externally; \\ supports dynamic updates \\ and reversibility.} \\
\midrule
Fine-Tuning 
& \makecell[l]{\xmark\ (FT, LoRA-FT) \\ \cmark\ (X-KDE, LTE)} 
& \makecell[l]{\xmark\ (FT, LoRA-FT) \\ \(\sim\)\ (X-KDE, LTE)} 
& \makecell[l]{High cost per update; \\ X-KDE enables reusable \\ batch edits.} \\
\midrule
Hypernetwork-Based 
& \cmark 
& \cmark 
& \makecell[l]{Meta-learned edits; \\ LiME supports low-latency, \\ language-conditioned updates.} \\
\bottomrule
\end{tabular}
}
\caption{Compatibility of MKE method families with sequential and batch editing. \cmark = supported; \(\sim\) = partial; \xmark = not supported.}
\label{tab:seq-batch}
\vspace{-2mm}
\end{table}

\paragraph{Sequential and Batch Editing} Realistic deployment requires models to support not only isolated edits but also sequential updates and batched revisions. Most current MKE methods are not evaluated in these conditions. Memory-based and hypernetwork approaches (e.g., ReMaKE, LiME) naturally support sequential updates via externalized edits, while parameter editors such as MEMIT suit batch updates. Instruction-tuned methods (e.g., X-KDE) enable reusable edit behavior in batches but incur higher training costs. Table \ref{tab:seq-batch} summarizes relative compatibility across method families.


\paragraph{Benchmark Fragmentation and Lack of Standardization}
The field still lacks unified evaluation protocols and benchmarks. While MLaKE and MZSRE are important steps forward, inconsistencies in construction, metrics, and coverage persist. Many rely on machine translation outputs, introducing noise. Low-resource and morphologically rich languages remain underrepresented. 

\paragraph{Domain-specific Editing} Current MKE research is almost entirely evaluated on general-purpose factual knowledge. The impact of editing in specialized domains such as biomedical, legal, or culturally grounded knowledge remains unexplored. Developing benchmarks and methods for domain-specific MKE would broaden applicability and expose new challenges in precision and transfer.

\paragraph{Scalability and Real-World Applicability}
Multilingual knowledge editing methods face key barriers to deployment. Memory-based approaches incur high storage or retrieval costs when scaling across languages. Fine-tuning is computationally expensive, especially when edits must generalize to variants. Parameter editors, though efficient, often lack robustness across diverse language families.

\subsection{Opportunities}


\paragraph{Development of Robust Multilingual Benchmarks}
Future benchmark design must move beyond machine translation and include high-quality, human-curated datasets covering low-resource languages and morphologically rich targets. Multi-hop reasoning tasks (e.g., MQA-KEAL) should be expanded to test compositional edit capabilities.

\paragraph{Language-Conditioned or Language-Aware Editing}
Early work such as LiME shows that explicitly conditioning edits on language identifiers can help mitigate representation anisotropy. Future models could leverage joint multilingual objectives or hierarchical representations to improve robustness across language pairs.

\paragraph{Instruction-Tuned and Task-Oriented Editing}
Instruction-tuned paradigms like X-KDE offer an attractive framework to model knowledge editing as a task. Continued exploration of instruction formats, task decomposition, and preference optimization could improve multilingual generalization.

\paragraph{Multilingual In-Context Learning and Adapter Fusion}
Combining prompting with adapter-based updates could enable lightweight multilingual editing with minimal model disruption.

\paragraph{Multi-source Editing} 
Most current MKE methods assume monolingual edits, with cross-lingual propagation only evaluated after the fact. A few approaches (e.g., LU-LAFNs, LiME) explicitly support multi-source editing by learning language-agnostic representations, allowing the same fact to be updated jointly across languages. This capability remains underexplored but represents an important opportunity for ensuring consistency and minimizing interference in multilingual models.

\paragraph{Cross-Model Editing and Transfer}
A largely unexplored avenue is whether knowledge edits can transfer across model families. Building model-agnostic edit representations could open new possibilities for scalable MKE.

\subsection{Connections to Adjacent Research Areas}

\paragraph{Hallucination mitigation}
Knowledge editing has been explored as a way to correct hallucinations by directly updating model representations, with benchmarks such as HalluBench~\cite{huang2025can} and TruthX~\cite{truthx}. However, evaluation remains English-only, and studies on multilingual hallucinations focus on measuring or reducing rates without using editing. Thus, the intersection of hallucination mitigation and MKE represents an exciting frontier for future research.
\paragraph{Mechanistic interpretability and cross-lingual transfer}
Many editing methods, particularly parameter-based ones, build on mechanistic interpretability, assuming factual knowledge can be localized in neural circuits. Yet little is known about whether such circuits align across languages or how misalignment affects propagation. Exploring this link could ground MKE theoretically and explain why some edits generalize cross-lingually while others remain language-specific.
Together, these connections show that MKE both informs and benefits from adjacent areas, though most current work remains monolingual.








\section*{Acknowledgments}
We thank the anonymous reviewers and the meta-reviewer for their thoughtful and constructive feedback. Their comments helped us refine the scope of this survey, improve the clarity of our presentation, and integrate additional discussions (e.g., model collapse, sequential and batch editing, and connections to adjacent areas). We are grateful for their input, which has significantly strengthened the final version of this paper.

\section*{Limitations}

While this survey provides a comprehensive overview of recent advances in multilingual knowledge editing (MKE), our work has several limitations:

\begin{itemize}

\item \textbf{Cross-paper comparability:} Our synthesis relies on reported results from prior studies. Due to variations in evaluation protocols, dataset versions, and metric definitions, exact comparisons across methods are not always possible. We therefore emphasize relative trends rather than absolute rankings.

\item \textbf{No new experimental validation:} We do not contribute new experimental results. Our findings on limitations, especially regarding low-resource and under-represented languages, are based on the coverage and reporting of the original papers.

\item \textbf{Selection completeness:} While we aimed to cover the most influential and representative papers across major method families, the rapidly evolving nature of multilingual knowledge editing means that relevant recent or niche approaches may have been unintentionally omitted.

\end{itemize}

\section*{Ethical Consideration}

This work is a survey of existing research on multilingual knowledge editing and does not involve any new data collection, user studies, or model deployment. We do not foresee any direct ethical concerns. All referenced works have been appropriately cited, and any datasets or models discussed were originally released under appropriate terms by their respective authors. AI tools were used to rephrase and improve exposition of sections of the paper.


\bibliography{custom}

\appendix

\begin{table*}[h!]
\centering
\begin{tabular}{lcccccc}
\hline
\textbf{Dataset/Method} & \textbf{Single-hop} & \textbf{Multi-hop} & \textbf{Reliability} & \textbf{Generality} & \textbf{Locality} & \textbf{Portability} \\
\hline
MLaKE               & \cmark & \cmark & \xmark & \xmark & \xmark & \xmark \\
mZsRE               & \xmark & \xmark & \cmark & \cmark & \cmark & \cmark \\
biZsRE              & \xmark & \xmark & \cmark & \cmark & \cmark & \cmark \\
Eva-KELLM           & \xmark & \xmark & \cmark & \cmark & \cmark & \cmark \\
BMIKE-53            & \xmark & \xmark & \cmark & \cmark & \cmark & \cmark \\
CROLIN-MQUAKE       & \cmark & \cmark & \xmark & \xmark & \xmark & \xmark \\
MQuAKE-AR           & \cmark & \cmark & \xmark & \xmark & \xmark & \xmark \\
MQA-AEval           & \cmark & \cmark & \xmark & \xmark & \xmark & \xmark \\
MKEB                & \xmark & \xmark & \cmark & \cmark & \cmark & \cmark \\
Counterfact (Catalan) & \xmark & \xmark & \cmark & \cmark & \cmark & \cmark \\
CKnowEdit           & \xmark & \xmark & \cmark & \cmark & \cmark & \cmark \\
BabelNet & \cmark & \cmark & \cmark & \cmark & \cmark & \cmark \\
\hline
\end{tabular}
\caption{Comparison of datasets/methods across reasoning type and evaluation dimensions.}
\label{tab:datasets_comparison}
\end{table*}

\section{Details on MKE Benchmarks}
\label{appendix:benchmarks}

A range of multilingual benchmarks have been proposed to evaluate MKE methods. Below, we summarize each dataset's design, construction process, and evaluation focus.

\textbf{MLaKE}~\cite{wei-etal-2025-mlake} is a benchmark comprising 4,072 multihop and 5,360 single-hop questions across five languages: English, Chinese, Japanese, French, and German. The dataset was constructed by aligning fact chains from Wikipedia, generating raw data via ChatGPT, and finalizing question-answer pairs through human verification. While MLaKE supports both single-hop and multihop evaluation, it lacks paraphrase queries and locality test samples.

\textbf{MzSRE}~\cite{wang-etal-2024-retrieval} is a machine-translated version of the ZsRE dataset in 12 languages: English, Czech, German, Dutch, Spanish, French, Portuguese, Russian, Thai, Turkish, Vietnamese, and Chinese. Translations were generated using Google Translate without human revision. Each sample is used to evaluate reliability, generality, locality, and portability.

\textbf{biZsRE}~\cite{wang-etal-2024-cross} is a bilingual version of ZsRE translated into Chinese. Training data was translated using GPT-3.5-turbo, while test samples were translated via GPT-4 and then validated by native Chinese speakers.

\textbf{Eva-KELLM}~\cite{wu2023evakellmnewbenchmarkevaluating} extends knowledge editing evaluation to full documents. Starting from the Counterfact dataset~\cite{ROME}, ChatGPT was used to generate documents corresponding to each edited fact. The dataset consists of 8,882 English documents and 6,930 Chinese documents.\footnote{Referred to as DocTer as of July 24, 2025}

\textbf{BMIKE-53}~\cite{nie2025bmike53investigatingcrosslingualknowledge}\footnote{Not publicly released as of May 15, 2025} evaluates cross-lingual in-context knowledge editing across 53 languages. It unifies ZsRE, Counterfact, and WikiFactDiff datasets, translating them into 52 languages using GPT-4o. Translation quality was verified qualitatively via manual inspection and quantitatively via BLEU scores on back-translated samples.

\textbf{CROLIN-MQUAKE}~\cite{khandelwal-etal-2024-cross}\footnote{Not publicly released as of May 15, 2025} is a translated version of the multihop model editing datasets MQuAKE-CF and MQuAKE-T~\cite{zhong-etal-2023-mquake}. It includes translations into German, Spanish, Chinese, Russian, Hindi, Bengali, and Swahili. Translations were reviewed by human experts and validated through back-translation BLEU scores.

\textbf{MQuAKE-AR and MQA-AEval}~\cite{ali-etal-2025-mqa} include Arabic translations of MQuAKE-CF and MQuAKE-T. The translation process combined LLM-based scoring (via ChatGPT) with manual refinement of low-quality samples. MQA-AEval is a manually curated Arabic benchmark for evaluating multihop edits.

\textbf{MKEB}~\cite{xie2024memlaenhancingmultilingualknowledge}\footnote{Not publicly available as of May 15, 2025} was constructed by querying Wikidata via SPARQL to retrieve factual triplets and chains. ChatGPT was used to generate corresponding questions, and translations into 12 languages (including English, Chinese, French, German, Japanese, Korean, and Portuguese) were performed via the Baidu API. No human revision was conducted.

\textbf{Counterfact (Catalan)}~\cite{MEMAT} is a translated version of the Counterfact dataset into Catalan. Translations were performed using the Aina English-Catalan translator and aligned using SimAlign~\cite{jalili-sabet-etal-2020-simalign}. Filters ensured sentence order consistency, and one round of human revision was conducted.

\textbf{CKnowEdit}~\cite{fang2024cknowedit} is a Chinese-centric benchmark designed to evaluate corrections of linguistic, factual, and logical errors. Unlike other datasets, it emphasizes language-specific and cultural relevance in Chinese.

\textbf{BabelEdits} ~\cite{green-etal-2025-babeledits} is a knowledge editing benchmarking comprising 60 languages. It combines multlingual synsets from babel net with marker based translation to ensure entity (subject) translation quality.  A comparison between the dataset is shown in Table \ref{tab:datasets_comparison}

    \label{fig:lu_lafns_metrics_comparison_qwen}

\end{document}